\documentclass[conference]{IEEEtran}
\usepackage{blindtext, graphicx}
\usepackage{mathtools}
\usepackage[utf8]{inputenc}
\usepackage[english]{babel}
\usepackage{comment}
\usepackage{csquotes}

\usepackage[backend=bibtex,style=ieee,sorting=ynt]{biblatex}
\DeclarePairedDelimiter{\ceil}{\lceil}{\rceil}

\ifCLASSINFOpdf
\else
\fi

\begin{document}
\title{Unsupervised Representation Learning of Structured Radio Communication Signals}

\author{\IEEEauthorblockN{Timothy J. O'Shea}
\IEEEauthorblockA{Bradley Department of Electrical\\ and Computer Engineering\\
Virginia Tech\\
Arlington, VA\\
http://www.oshearesearch.com}
\and
\IEEEauthorblockN{Johnathan Corgan}
\IEEEauthorblockA{Corgan Labs\\
San Jose, CA\\
http://corganlabs.com/}
\and
\IEEEauthorblockN{T. Charles Clancy}
\IEEEauthorblockA{Bradley Department of Electrical\\ and Computer Engineering\\
Virginia Tech\\
Arlington, VA\\
http://www.stochasticresearch.com/}
}

\maketitle

\begin{abstract}
We explore unsupervised representation learning of radio communication signals in raw sampled time series representation.  We demonstrate that we can learn modulation basis functions using convolutional autoencoders and visually recognize their relationship to the analytic bases used in digital communications.  We also propose and evaluate quantitative metrics for quality of encoding using domain relevant performance metrics.
\end{abstract}


\IEEEpeerreviewmaketitle

\section{Introduction}

Radio signals are all around us and serve as a key enabler for both communications and sensing as our world grows increasingly reliant on both in a heavily interconnected and automated world.  Much effort has gone into expert system design and optimization for both radio and radar systems over the past 75 years considering exactly how to represent, shape, adapt, and recover these signals through a lossy, non-linear, distorted, and often interference heavy channel environment. 
Meanwhile, in recent years, heavily expert-tuned basis functions such as Gabor filters in the vision domain have been largely discarded due to the speed at which they can be naively learned and adapted using feature learning approaches in deep neural networks.  

Here we explore making the same transition from using relatively simple expert-designed representation and coding to using emergent, learned encoding.  We expect to better optimize for channel capacity, to be able to translate information to and from channel and compact representations, and to better reason about what kind of information is in the radio spectrum--allowing less-supervised classification, anomaly detection, and numerous other applications.

This paper provides the first step towards that goal by demonstrating that common radio communications signal bases emerge relatively easily using existing unsupervised learning methods.  We outline a number of techniques which enable this to work to provide insight for continued investigation into this domain.  This work extends promising prior supervised feature learning work in the domain we have already begun in \cite{convmodrec}.

\subsection{Basis Functions for Radio Data}

\begin{figure}[ht!]
  \centering
      \includegraphics[width=0.5\textwidth]{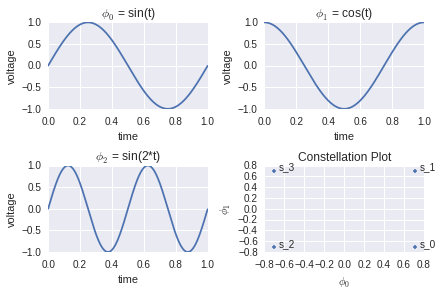}
        \caption{Example Radio Communications Basis Functions}
    \label{fig:basis}
\end{figure}

Widely used single-carrier radio signal time series modulations schemes today still use a relatively simple set of supporting basis functions to modulate information into the radio spectrum.  Digital modulations typically use a set of sine wave basis functions with orthogonal or pseudo-orthogonal properties in phase, amplitude, and/or frequency.  Information bits can then be used to map a symbol value $s_i$ to a location in this space $\phi_j, \phi_k, ...$.  In figure \ref{fig:basis} we show three potential basis functions where $\phi_0$ and $\phi_1$ form phase-orthogonal bases used in Phase Shift Keying (PSK) and Quadrature Amplitude Modulation (QAM), while $\phi_0$ and $\phi_2$ show frequency-orthogonal bases used in Frequency Shit Keying (FSK).  In the final figure of \ref{fig:basis} we show a common mapping of constellation points into this space as typically used in Quadrature Phase Shift Keying (QPSK) where each symbol value encodes two bits of information.

Digital modulation theory in communications is a rich subject explored in much greater depth in numerous great texts such as \cite{sklar2001digital}.

\subsection{Radio Signal Structure}

\begin{figure}[ht!]
  \centering
      \includegraphics[width=0.5\textwidth]{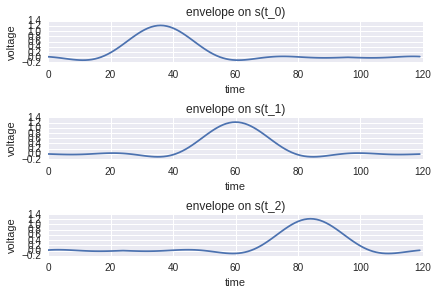}
        \caption{Discrete Symbols Envelopes in Time}
    \label{fig:rrc}
\end{figure}

Once basis functions have been selected, data to transmit is divided into symbols and each symbol period for transmission occupies a sequential time slot.  To avoid creating wideband signal energy associated with rapid transitions in symbols, a pulse shaping envelope such as a root-raised cosine or sinc filter is typically used to provide a smoothed transition between discrete symbol values in adjacent time-slots \cite{rrc}.  Three such adjacent symbol time slots can be seen in figure \ref{fig:rrc}.   Ultimately a sequence of pulse shaped symbols with different values are summed together to form the transmit signal time-series, $s(t)$.

\subsection{Radio Channel Effects}

The transmitted signal, $s(t)$, passes through a number of channel effects over the air before being received as $r(t)$ at the receiver.  This includes time-delay, time-scaling, phase rotation, frequency offset, additive thermal noise, and channel impulse responses being convolved with the signal, all as random unknown time-varying processes.  A closed form of all these effects might take the form of something roughly like this:

\begin{equation} \label{eq:realistic}
r(t) = e^{j*n_{Lo}(t)} \int_{\tau=0}^{\tau_0} s(n_{Clk}(t-\tau))h(\tau) + n_{Add}(t) 
\end{equation}

This significantly complicates the data representation from its original straightforward encoding at the transmitter when considering the effects of wireless channels as they exist in the real world.

\section{Learning from Radio Signals}

We focus initially on attempting to learn symbol basis functions from existing modulation schemes in wide use today.  We focus on Quadrature Phase-Shift Keying (QPSK) and Gaussian Binary Frequency Shift Keying (GFSK) as our modulation of interest in this work and hope to demonstrate learning the analytical basis functions for these naively.

\subsection{Building a Dataset}

We leverage the dataset from \cite{convmodrec} and focus on learning only a single modulation basis set at a time in this work.   This dataset includes the QPSK and GFSK modulations passed through realistic, but relatively benign wireless channels, sampled in 88 complex-valued sample times per training example.

\subsection{Unsupervised Learning}

Autoencoders \cite{autoencoders} have become a powerful and widely used unsupervised learning tool.  We review the autoencoder and several relevant improvements on the autoencoder with application to this domain which we leverage.

\subsubsection{Autoencoder Architectures}
Autoencoders (AE) learn an intermediate, possibly lower dimensional encoding of an input by using reconstruction cost as their optimization criteria, typically attempting to minimize Mean Squared-Error (MSE).  They consist of an encoder which encodes raw inputs into a lower-dimension hidden sparse representation, and a decoder which reconstructs an estimate for the input vector as the output.

A number of improvements have been made on autoencoders which we leverage below.

\subsubsection{Denoising Autoencoders}
By introducing noise into the input of an AE training, but evaluating its reconstruction of the unmodified input, Denoising Autoencoders \cite{dnae} perform an additional input noise regularization effect which is extremely well suited in the communications domain where we always have additive Gaussian thermal noise applied to our input vectors.

\subsubsection{Convolutional Autoencoders}
Convolutional Autoencoders \cite{convae} are simply autoencoders leveraging convolutional weight configurations in their encoder and decoder stages.  By leveraging convolutional layers rather than fully connected layers, we force time-shift invariance learning in our features and reduce the number of parameters required to fit.  Since our channel model involves random time shifting of the input signal, this is an important property to the radio application domain which we feel is extremely well suited for this task.

\subsubsection{Regularization}

We leverage heavy $L_2 = {\left \| \mathbf{W}  \right \|}_2$ weight regularization and $L_1 = {\left \| \mathbf{h}  \right \|}_1$ activity regularization in our AE to attempt to force it to learn orthogonal basis functions with minimal energy.   \cite{l1l2reg}
Strong $L_1$ activation regularization is especially important in the narrow hidden layer representation between encoder and decoder where we would like to learn a maximally sparse compact basis representation of the signal through symbols of interest occurring at specific times.
Dropout \cite{dropout} is also used as a form of regularization between intermediate layers, forcing the network to leverage all available weight bases to span the representation space.

\subsection{Test Neural Network Architecture}

Our goal in this effort was to obtain a minimum complexity network which allows us to convincingly reconstruct the signals of interest with a significant amount of information compression.  By using convolutional layers with only one or two filters, we seek to achieve a maximally matched small set of time-basis filters with some equivalence to the expert features used to construct the signal.  Dense layers with non-linear activations then sit in between these to provide some estimation of the logic for what the representation and reconstruction should be for those basis filters occurring at different times.
The basic network architecture is shown below in figure \ref{fig:arch}.

\begin{figure}[ht!]
  \centering
      \includegraphics[width=0.5\textwidth]{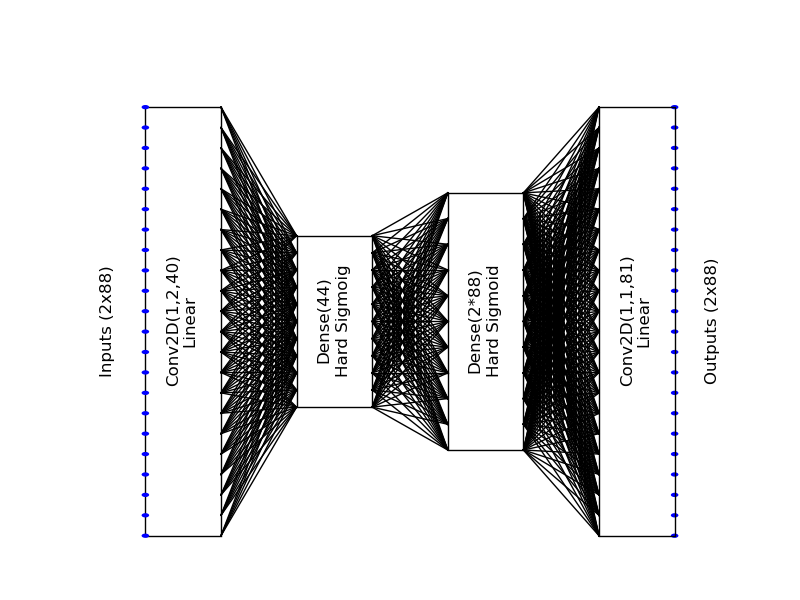}
        \caption{Convolutional Autoencoder Architecture Used}
    \label{fig:arch}
\end{figure}

\subsection{Evaluation Methods for Reconstruction}

For the scope of this work we use MSE as our reconstruction metric for optimization.  We seek to evaluate reconstructed signals from BER and SNR, but plan to defer this for later work in the interest of space.

\subsection{Visual Inspection of Learned Representations}

\begin{figure}[ht!]
  \centering
      \includegraphics[width=0.5\textwidth]{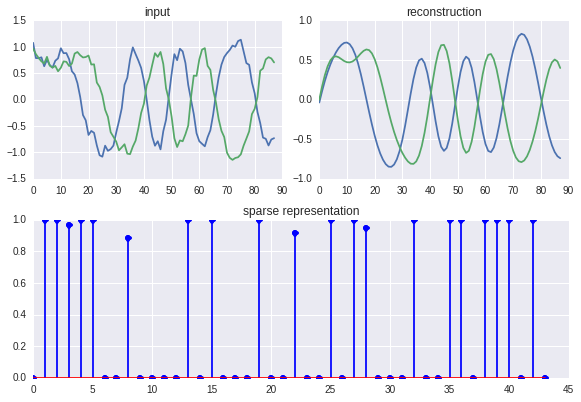}
        \caption{QPSK Reconstruction 1 through Conv-AE}
    \label{fig:recon1}
\end{figure}

\begin{figure}[ht!]
  \centering
      \includegraphics[width=0.5\textwidth]{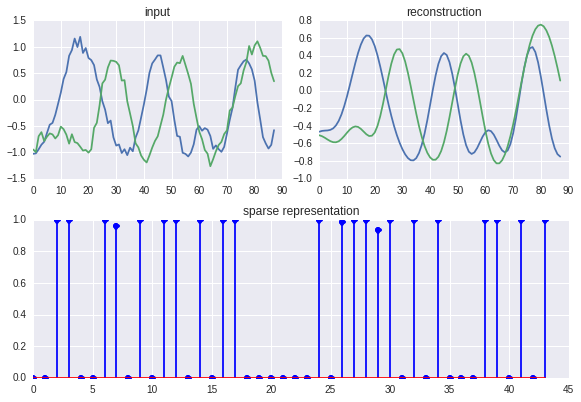}
        \caption{QPSK Reconstruction 2 through Conv-AE}
    \label{fig:recon2}
\end{figure}

Given a relatively informed view of what a smooth band-limited QPSK signal looks like in reality, visual inspection of the reconstruction vs the noisy input signal is an important way to consider the quality of the representation and reconstruction we have learned.  The sparse representation is especially interesting as by selecting hard-sigmoid dense layer activations we have effectively forced the network to learn a binary representation of the continuous signal.   Ideally there exists a direct GF(2) relationship between the encoded bits and the coded symbol bits of interest here.  Figures \ref{fig:recon1} and \ref{fig:recon2} illustrate this reconstruction and sparse binary representation learned.


\begin{figure}[ht!]
  \centering
      \includegraphics[width=0.5\textwidth]{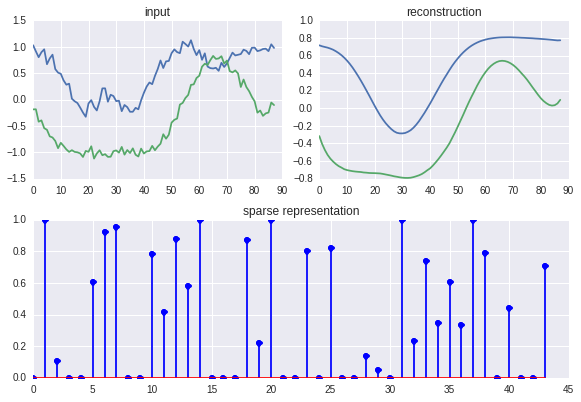}
        \caption{GFSK Reconstruction 1 through Conv-AE}
    \label{fig:fsk1}
\end{figure}

For GFSK, we show reconstructions and sparse representations in figure \ref{fig:fsk1}.  In this case, the AE architecture converges even faster to a low reconstruction error, but unfortunately the sparse representations are not saturated into discrete values as was the case for the constant modulus signal.

\section{Results}

We consider the significance of these results below in the context of the network complexity required for representation and the compression ratio obtained.

\subsection{Learned Network Parameters}

We use Adam \cite{kingma2014adam} (a momentum method of SGD) to train our network parameters as implemented in the Keras \cite{keras} library.  Evaluating our weight complexity, we have two 2D convolutional layers, 2x1x1x40 and 1x1x1x81, making a total of only 161 parameters learned in these layers to fit the translation invariant filter features which form the primary input and output for our network.  The Dense layers which provide mappings from occurrences of these filter weights to a sparse code and back to a wide representation, consist of weight matrices of 516x44 and 44x176 respectively, making a total of 30448 dense floating point weight values.

Training is relatively trivial with this size network and dataset, we converge on a solution after about 2 minutes of training, 25 epochs on 20,000 training examples using a Titan X GPU.

\begin{figure}[ht!]
  \centering
      \includegraphics[width=0.5\textwidth]{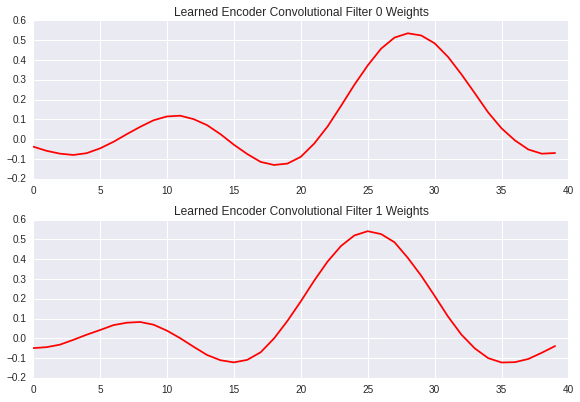}
        \caption{QPSK Encoder Convolutional Weights}
    \label{fig:conv_w1}
\end{figure}

\begin{figure}[ht!]
  \centering
      \includegraphics[width=0.5\textwidth]{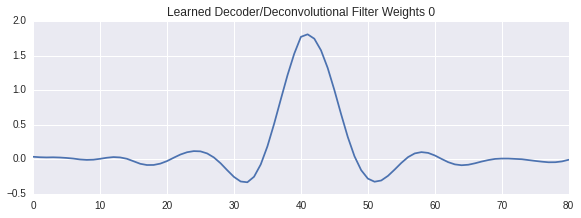}
        \caption{QPSK Decoder Convolutional Weights}
    \label{fig:conv_w2}
\end{figure}

In figure \ref{fig:conv_w1} we show the learned convolutional weight vectors in the encoder first layer.  We can clearly see a sinusoid occurs at varying time offests to form detections, and a second sinusoid at double the frequency, both with some minimal pulse shaping apparent on them.

In the decoder convolutional weight vector in figure \ref{fig:conv_w2} we can clearly see the pulse shaping filter shape emerge in the 

\begin{figure}[ht!]
  \centering
      \includegraphics[width=0.5\textwidth]{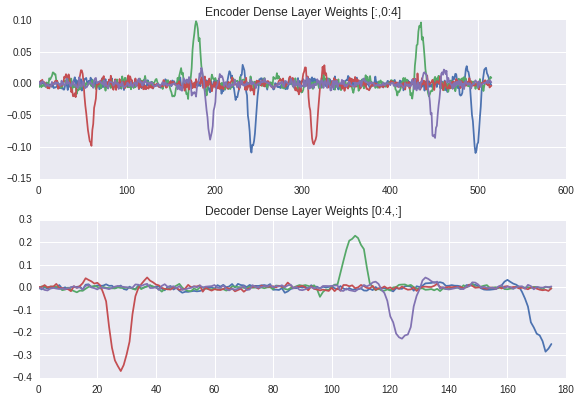}
        \caption{First Four Sparse Representation Dense Weights}
    \label{fig:dense_w}
\end{figure}

In figure \ref{fig:dense_w} we display the learned dense layer weights mappings of various symbol value and offset areas as represented by the convolutional filters.   It is important to note that the 1x516 input is a linearized dimension of zero-padded I and Q inputs through two separate filters (2x2x129).  We see that a single sparse hidden layer value equates to two pulses representing sinusoidal convolutional filter occurrences in time in the I and the Q channel, with roughly a sinc or root raised cosine window roll-off visibly represented in this time-scale.

\subsection{Radio Signal Representation Complexity}

To measure the compression we have achieved, we compare the effective number of bits required to represent the dynamic range in the input and output continuous signal domains with that of the number of bits required to store the signal in the hidden layer. \cite{infotheory}

If we consider that our input signal contains roughly 20dB of signal-to-noise ratio, we can approximate the number of bits required to represent each continuous value as follows. 

\begin{equation}
    N_{eff} = \ceil{ \frac{20dB - 1.76}{6.02} } = 4 \text{ bits}
\end{equation}

Given that we have 88*2 inputs of 4 bit resolution, compressed to 44 intermediate binary values, we get a compression ratio of \textbf{16x} = 88*2*4/44.

Given that we are learning roughly 4 to 5 symbols per example, with 4 samples per symbol, this equates to something like 10 bits being the most compact possible form of data-information representation.   However in the current encoder, we are also encoding timing offset information, phase error, and generally all channel information needed to reconstruct the data symbols in their specific arrival mode.   Given this is on the order of \textbf{4x} the most compact representation possible for the data symbols alone, this is not a bad starting point.

\section{Conclusions}

We are able to obtain relatively good compression with autoencoders for radio communications signals, however these must encode both the data bits and the channel state information which limits attainable compression.

Hard-sigmoid activations surrounding the hidden layer, for constant modulus modulations, seem effective in saturating representation into compact binary vectors, allowing us to encode 88 x 64 bit complex values into 44 bits of information without significant degradation.

Convolutional autoencoders are well suited for reducing parameter space, forcing time-invariance features, and forming a compact front-end for radio data.  We look forward to evaluating more quantitative metrics on reconstructed data, evaluating additional multi-level binary or hard-sigmoid representation for multi-level non-constant-modulus signals and investigating the use of attention models to remove channel variance from compact data representation requirements.

\section*{Acknowledgments}

The authors would like to thank the Bradley Department of Electrical and Computer Engineering at the Virginia Polytechnic Institute and State University, the Hume Center, and DARPA all for their generous support in this work.

This research was developed with funding from the Defense Advanced Research Projects Agency's (DARPA) MTO Office under grant HR0011-16-1-0002. The views, opinions, and/or findings expressed are those of the author and should not be interpreted as representing the official views or policies of the Department of Defense or the U.S. Government.

\nocite{clancy2007applications}

\printbibliography

\end{document}